\documentclass[preprint,number,12pt]{elsarticle}  % review al posto di preprint,number etc.

\usepackage{amssymb}
\usepackage{amsmath,amssymb,amsfonts}
\usepackage{amsthm}
\usepackage{color}
\usepackage{lineno,hyperref}
\usepackage{geometry}
\newgeometry{top=0.75in, bottom=1.2in, left=0.66in, right=0.66in}
\modulolinenumbers[5]
\usepackage{caption}

\journal{ATT'24: October 19, 2024, Santiago de Compostela, Spain}

\RequirePackage{algorithm}
\RequirePackage{algorithmic}
\usepackage{subcaption}
\usepackage{url}
\usepackage{comment}
\usepackage{siunitx}
\usepackage{lineno}
\usepackage{booktabs}

\usepackage[UKenglish]{babel}

\definecolor{mylimegreen}{RGB}{30, 200, 30}

\newcommand{\mfa}[1]{\textcolor{black}{#1}}
\newcommand{\mtc}[1]{\textcolor{black}{#1}}

\begin{document}

\begin{frontmatter}

\title{A Distributed Approach to Autonomous Intersection\\ Management via Multi-Agent Reinforcement Learning} %\tnoteref{mytitlenote}
%\tnotetext[mytitlenote]{Fully documented templates are available in the elsarticle package on \href{http://www.ctan.org/tex-archive/macros/latex/contrib/elsarticle}{CTAN}.}

%% Group authors per affiliation:
\author{Matteo Cederle\fnref{myfootnote}, Marco Fabris and Gian Antonio Susto}
\address{Dept. of Information Eng., Univ. of Padova, via Gradenigo 6/B, Padua, 35131, Italy.}
\fntext[myfootnote]{M.~Cederle, M.~Fabris and G.A.~Susto are with the Department of Information Engineering at the University of Padova (UNIPD - DEI), Padova, 35131, Italy (e-mails: matteo.cederle@phd.unipd.it, marco.fabris.1@unipd.it, gianantonio.susto@unipd.it).}

%% or include affiliations in footnotes:
%\author[mymainaddress,mysecondaryaddress]{Marco Fabris}
%\ead[url]{www.elsevier.com}
%
%\author[mysecondaryaddress]{Marco Fabris\corref{mycorrespondingauthor}}
%\cortext[mycorrespondingauthor]{Corresponding author}
%\ead{support@elsevier.com}
%
%\address[mymainaddress]{1600 John F Kennedy Boulevard, Philadelphia}
%\address[mysecondaryaddress]{360 Park Avenue South, New York}
	
\begin{abstract}
Autonomous intersection management (AIM) poses significant challenges due to the intricate nature of real-world traffic scenarios and the need for a highly expensive centralised server in charge of simultaneously controlling all the vehicles. This study addresses such issues by proposing a novel distributed approach to AIM utilizing multi-agent reinforcement learning (MARL). We show that by leveraging the 3D surround view technology for advanced assistance systems, autonomous vehicles can accurately navigate intersection scenarios without needing any centralised controller. The contributions of this paper thus include a MARL-based algorithm for the autonomous management of a 4-way intersection and also the introduction of a new strategy called \textit{prioritised scenario replay} for improved training efficacy. 
We validate our approach as an innovative alternative to conventional centralised AIM techniques, ensuring the full reproducibility of our results. Specifically, experiments conducted in virtual environments using the SMARTS platform highlight its superiority over benchmarks across various metrics.
\end{abstract}

\begin{keyword}
%\texttt{elsarticle.cls}\sep \LaTeX\sep Elsevier \sep template
Autonomous Intersection Management \sep
Connected Autonomous Vehicles \sep
DQN \sep
Multi-Agent Reinforcement Learning \sep
Reinforcement Learning \sep
Smart Mobility \sep
Traffic Scenarios
\end{keyword}

\end{frontmatter}

%\linenumbers

%\linenumbers

\section{Introduction}
\label{sec:intro}

\mfa{Connected autonomous vehicles (CAVs) represent a groundbreaking advancement in transportation, poised to revolutionize mobility by redefining commuting, parking, travel, and urban interaction~\cite{kopelias2020connected,savithramma2022smart}. Equipped with advanced sensors and AI systems, CAVs navigate roads with precision, reducing accidents caused by human error~\cite{papadoulis2019evaluating,szenasi2021statistical}. This enhanced safety feature not only saves lives but also makes mobility more accessible and inclusive for individuals with disabilities or vulnerabilities~\cite{brewer2018understanding,dicianno2021systematic}. On a societal level, CAVs optimize traffic flow and minimize congestion, reducing travel times and improving overall efficiency and stability~\cite{hyldmar2019fleet,talebpour2016influence}. Additionally, CAVs promise environmental sustainability by integrating electric and hybrid propulsion systems, significantly reducing greenhouse gas emissions~\cite{conlon2019greenhouse,taiebat2018review}.}

Lately, significant strides in the development of CAVs have been made, largely attributed to the utilization of multi-agent reinforcement learning (MARL)~\cite{schimdt2022introduction} within the framework of smart mobility,
showing promise in addressing autonomous intersection management (AIM)~\cite{dresner2004multiagent}. As it is widely believed that the resolution of AIM is pivotal to overcome in order to advance the adoption of CAVs, this control problem constitutes the primary focus for our study. 

A vast literature exists on AIM. The research in this field spans multiple fronts, each leveraging distinct methodologies to address the challenges of optimizing traffic flow and ensuring safety in dynamic urban environments.
By employing reinforcement learning (RL), AIM systems can effectively learn and adapt intersection control strategies in response to changing traffic conditions~\cite{karthikeyan2022autonomous,ayeelyan2022advantage}. These systems typically comprise priority assignment models, intersection control model learning, and safe brake control mechanisms. Experimental simulations demonstrate the superiority of RL-inspired AIM approaches over traditional methods, showcasing enhanced efficiency and safety.
Graph neural networks (GNNs) have also garnered attention for their potential in AIM~\cite{klimke2022cooperative,klimke2022enhanced}. Leveraging RL algorithms, GNNs optimize traffic flow at intersections by jointly planning for multiple vehicles. These models encode scene representations efficiently, providing individual outputs for all involved vehicles.
Game theory serves then as a foundational framework for MARL approaches in AIM. Indeed, game-theoretic models facilitate safe and adaptive decision-making for CAVs at intersections~\cite{li2023safe,liu2023potential}. By considering the diverse behaviors of interacting vehicles, these algorithms ensure flexibility and adaptability, thus enhancing autonomous vehicle performances in challenging scenarios.
Finally, recursive neural networks (RNNs) integrated in the MARL framework represent an interesting approach in AIM research %, with the objective of learning complex traffic dynamics and optimizing vehicle speed control
\mfa{to learn complex traffic dynamics and optimize vehicle speed control}~\cite{guillenperez2022multiagent}. %By harnessing the power of deep RL within a multi-agent framework through the integration of deep neural networks~\cite{liu2017survey}, MADRL-based AIM systems autonomously 

Despite the advancements in AIM techniques, their implementation still faces important challenges. %The main obstacle is represented by the need for a highly expensive centralised server which has to be positioned in the proximity of the intersection, in order to simultaneously control all the vehicles.  
\mfa{One of the main obstacles is represented by the need for an expensive centralised server which has to be positioned in the proximity of the intersection, in order to simultaneously control all the vehicles.} Moreover, the vehicles should continuously send their local information to this centralised controller, which will then gather and elaborate the data coming from all the road users, before sending back to each vehicle a velocity or acceleration command. 
Given the complexity and high demands of this technological framework, the integration of AIM devices into existing transportation infrastructures still requires many years of extensive research and testing. In this direction, we devise an alternative distributed approach based on the 3D surround view technology for advanced assistance systems~\cite{gao20173}. As shown in the sequel, such a method allows the reconstruction of a 360$^{\circ}$ scene centered around each CAV, which is useful to recover the information required for each agent involved in the proposed MARL-based technique. This, in turn, grants to effectively carry out AIM in a decentralised fashion, exploiting sensors that are currently available on the market, without the need for the centralised infrastructure described in the previous lines. More precisely, the contributions of this paper are multiple.
%\forse{Despite the advancements in AIM techniques, their implementation still faces challenges. One significant obstacle is the complexity of real-world traffic scenarios, which often accounts for diverse and unpredictable interactions among various road users, possibly including non-autonomous vehicles, pedestrians and cyclists. Additionally, ensuring the safety and reliability of AIM systems in dynamic environments remains a paramount concern. For this reason, the integration of AIM devices into existing transportation infrastructures still demands extensive testing and validation.
%In this direction, we devise an alternative approach based on the 3D surround view technology for advanced assistance systems~\cite{gao20173}. As demonstrated in the sequel, such a method allows the reconstruction of a 360$^{\circ}$ scene centered around each CAV, which is useful to recover the information required for each agent involved in the proposed MARL-based technique. This, in turn, grants to effectively carry out AIM in a decentralised fashion exploiting cheap sensors that are currently available on the market.
%More precisely, the contributions of this paper are multiple.}
\begin{itemize}
    \item As mentioned above, we offer a new distributed strategy that represents a competitive and realistic alternative to the classical centralised AIM techniques.
    \item Relying on \textit{self-play} \cite{chakraborty2014multiagent} and drawing inspiration from \textit{prioritised experience replay} \cite{schaul2015prioritized} to improve training efficacy, we develop a MARL-based algorithm capable of tackling and solving a 4-way intersection by means of the SMARTS platform \cite{zhou2021smarts}. 
	\item Our strategy outperforms a number of well-established benchmarks, which typically leverage traffic light regulation in function of travel time, waiting time and average speed.
	\item Last but not least, we guarantee full reproducibility\footnote{The Python code of our work can be found at \url{https://github.com/mcederle99/MAD4QN-PS}. The authors want to stress the fact that reproducibility represents a crucial issue within this field of research. %Unfortunately, in many interesting related works we did not find reproducible code \cite{karthikeyan2022autonomous, ayeelyan2022advantage, guillenperez2022multiagent}.
    }
    %Indeed, this aspect is often heavily neglected in almost all the related works.
    of the code that is used for the generation of the virtual experiments shown in this manuscript. 
\end{itemize}

The remainder of this paper unfolds as follows. The preliminaries for this study are yielded in Section~\ref{sec:backgr}; whereas, Section \ref{sec:method} provides the core of our contribution, namely the~\textit{multi-agent decentralised dueling double deep q-networks algorithm with prioritised scenario replay (MAD4QN-PS)}. This innovative method is then tested and validated through several virtual experiments, as illustrated in Section~\ref{sec:experiments}. Finally, Section~\ref{sec:conclusions} draws the conclusions for the present investigation, proposing future developments.

\textit{Notation}:
The sets of natural and positive (zero included) real numbers are denoted by $\mathbb N$ and $\mathbb R_{0}^{+}$, respectively. Given a random variable (r.v.) $Y$, its probability mass function is denoted by $P[Y=y]$, whereas $P[Y=y ~|~ Z=z]$ indicates the probability mass function of $Y$ conditioned to the observation of a r.v. $Z$. Moreover, the expected value of a r.v. $Y$ is denoted by $\mathbb E[Y]$. % (or $\mathbb{E}_z[Y(z)]$, whenever the argument $Y$ depends on the values $z$ of a hidden r.v. $Z$).

\section{Theoretical background}
\label{sec:backgr}

\subsection{Basic notions of reinforcement learning}
\label{subsec:rl}
RL is a machine learning paradigm in which an \textit{agent} learns to solve a \textit{task} by iteratively interacting with its \textit{environment}. Solving the task means maximising the cumulative rewards obtained over time. A generic RL problem is formalised by the concept of \textit{Markov decision process (MDP)} \cite{sutton2018reinforcement}, which is a tuple composed by five elements: ${\left\langle \mathcal{S}, \mathcal{A}, \mathcal{P}, \mathcal{R}, \gamma \right\rangle}$. $\mathcal{S}$ and $\mathcal{A}$ are two generic sets, representing the \textit{state} and \textit{action space} respectively. ${\mathcal{P}\left(s, a, s'\right) = P\left[S_{t+1}=s' \mid S_t=s, A_t=a\right]}$ is the \textit{state transition probability function}, in charge of updating the environment to a new state $s'\in \mathcal{S}$ at each step, based on the previous state $s\in \mathcal{S}$ and the action $a\in \mathcal{A}$ performed by the agent. Moreover, the \textit{reward function} $\mathcal{R}(s, a, s'): \mathcal{S} \times \mathcal{A} \times \mathcal{S} \rightarrow \mathbb{R}$ is used to measure the quality of each transition, while $\gamma \in [0,1)$ denotes a \textit{discount factor}, used to compute the \textit{cumulative reward} at time $t$, i.e. the \textit{return} $G_t = \sum_{k=0}^{\infty}\gamma^k r_{t+k+1}$. The agent decides which action to take at each iteration exploiting its \textit{policy}, a function that maps any state to the probability of selecting each possible action:
\begin{equation}
    \pi(a|s) = P[A_t = a \mid S_t = s], \hspace{0.3cm} \forall
    a \in \mathcal{A}.
\end{equation}
Solving a RL problem means finding an optimal policy $\pi^*$. One criterion that is usually adopted to find $\pi^*$ consists in the maximization of the \textit{state-action value function} $Q_\pi(s,a)$, i.e. the expected return starting from state $s\in \mathcal{S}$, taking action $a\in \mathcal{A}$, and thereafter following policy $\pi$:
\begin{equation}
Q_\pi(s,a) = \mathbb{E}_{\pi} \left[G_t \mid S_t = s, A_t = a\right] .  
\end{equation}
Consequently, given the state-action value function, the optimal policy is defined as $\pi^* = \text{arg}\max_\pi Q_\pi(s,a)$. There is therefore an inherent relation between $\pi^*$ and the optimal state-action value function.

Finally, two other important quantities which will be used in the proceeding of this article are the \textit{state value function} and the \textit{advantage function}. The former is defined as the expected return starting from state $s\in \mathcal{S}$ and then following policy $\pi$:
\begin{equation} \label{eq:vfunc}
V_\pi(s) = \mathbb{E}_{\pi} \left[G_t \mid S_t = s\right] .  
\end{equation}
The latter instead is used to give a relative measure of importance to each action for a particular state, and it is defined starting from $Q_\pi(s,a)$ and $V_\pi(s)$:
\begin{equation} \label{eq:advfunc}
A_\pi(s,a) = Q_\pi(s,a) - V_\pi(s) .  
\end{equation}

\subsection{Q-Learning and Deep Q-Networks}
\label{subsec:qldqn}
To compute the optimal state-action value function we could theoretically exploit the recursive \textit{Bellman Optimality Equation} \cite{bellman1957markovian}:
\begin{equation} \label{eq:bellman}
    Q^*(s,a) = \mathbb{E} \left[r_{t+1} + \gamma \max_{a'}Q^*(S_{t+1},a') \mid S_t = s, A_t = a\right],
\end{equation}
however due to the curse of dimensionality and the need for perfect statistical information to compute the closed-form solution, it is necessary to resort to iterative learning strategies even to solve simple RL problems. The most common algorithm used in literature is \textit{Q-Learning} \cite{watkins1989learning}, where the state-action value function is represented by a table, which is iteratively updated at each step through an approximation of (\ref{eq:bellman}):
\begin{equation} \label{eq:updateql}
    Q_{t+1}(s_t,a_t) \leftarrow \hspace{0.1cm} Q_t(s_t,a_t) + \alpha(r_{t+1} + \gamma \max_{a'}Q_t(s_{t+1},a') - Q_t(s_t,a_t)) ,
\end{equation}
where $\alpha>0$ is called the \textit{step-size parameter}. The policy derived from the state-action value function is usually the $\varepsilon$-greedy policy, suitable to balance the trade-off between \textit{exploration} and \textit{exploitation}~\cite{sutton2018reinforcement}:
\begin{equation}
    \pi(a|s) = \begin{cases}
        \text{arg}\max_{a} Q(s,a) \hspace{0.5cm} &\text{with probability $1-\varepsilon$} \\
        \text{random action } a \in \mathcal{A} &\text{with probability $\varepsilon$}
    \end{cases}
\end{equation}
Tabular Q-Learning works well for simple tasks, but the problem rapidly becomes intractable when the state space becomes very large or even continuous. For this reason state-of-the-art RL algorithms employ function approximators, such as \textit{neural networks} (NNs), to solve realistic and complex problems. One of the first yet more used deep RL algorithms is \textit{Deep Q-Networks} \cite{mnih2015human}, which approximates the state-action value function through a NN, $Q(s,a;\theta)$. A \textit{replay memory} is used to store the transition tuples $(s,a,r,s')$. Finally, the parameters $\theta$ of the Q-Network are optimised by \textit{sampling batches} $\mathcal{B}$ of transitions from the replay memory and minimizing a mean squared error loss derived from (\ref{eq:updateql}):
\begin{equation} \label{eq:loss}
    \mathcal{L}(\theta) = \cfrac{1}{|\mathcal{B}|}\sum_{i\in \mathcal{B}} [(r_i+\gamma \max_{a'}Q(s_i',a';\bar{\theta}) - Q(s_i,a_i;\theta))^2] ,
\end{equation}
where $\bar{\theta}$ represent the parameters of a target network, which are periodically duplicated from $\theta$ and maintained unchanged for a predefined number of iterations.

\subsection{Multi-agent reinforcement learning}
\label{subsec:marl}
MARL expands upon traditional RL by incorporating \textit{multiple} agents, each making decisions in an environment where their actions influence both the immediate rewards and the observations of other agents. In its most general definition, a MARL problem is formalised as a \textit{partially observable stochastic game} (POSG), in which each agent has its own action space and reward function. Moreover, the partial observability derives from the fact that the agents do not perceive the global state, but just local observations, which carry incomplete information about the environment \cite{shapley1953stochastic}.

MARL algorithms can be categorised depending on the type of information available to the agents during training and execution: in centralised training and execution (CTCE), the learning of the policies as well as the policies themselves use some type of structure that is centrally shared between the agents. On the other hand, in decentralised training and execution (DTDE), the agents are fully independent and do not rely on centrally shared mechanisms. Finally, the centralised training and decentralised execution paradigm (CTDE) is in between the first two, exploiting centralised training to learn the policies, while the execution of the policies themselves is designed to be decentralised~\cite{albrecht2023multi}.

\section{Multi-Agent Decentralised Dueling Double Deep Q-Networks with Prioritized Scenario replay}
\label{sec:method}
In this section we present our novel method based on MARL, called Multi-Agent Decentralised Dueling Double Deep Q-Networks with Prioritized Scenario replay (MAD4QN-PS). We begin by detailing how the system is modelled, and then we describe the original learning procedure that we implement in order to train agents through \textit{self-play} \cite{chakraborty2014multiagent}. Finally, we shall introduce the \textit{prioritised scenario replay} pipeline that is implemented to speed up training.

\begin{figure}[b]
 \centering
  \includegraphics[width=5cm]{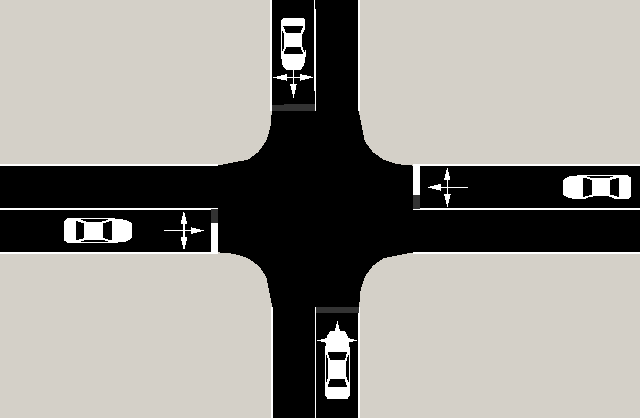}
  \caption{Visual representation of the 4-way 1-lane intersection environment considered for the experiments (image created through SUMO simulator \cite{SUMO2018}).}
  \label{fig:scenario}
\end{figure}

\subsection{System modelling and design}
\label{subsec:sysmod}
The environment in which the agents live consists of a 4-way 1-lane intersection, with three different turning intentions available to each vehicle, as shown in Figure \ref{fig:scenario}.

Recalling Section \ref{subsec:marl}, we formalize the problem as a POSG, which can be seen as a multi-agent extension to MDPs. For this reason, we shall define for each agent the observation space, the action space and the reward function.

The information retrieved by each vehicle at every time step consists of a local RGB bird-eye view image with the vehicle at the center. As already discussed in Section \ref{sec:intro}, this type of data is already recoverable from sensors with modern technology, thus making such a configuration extremely interesting from an application point of view. Moreover, the final observation passed to the agent is represented by a stack of $n\in \mathbb{N}$ consecutive frames, thus allowing the algorithm to capture temporal dependencies and understand how the environment is changing over time.

The action space of each agent instead is discrete and it contains $m\in \mathbb{N}$ velocity commands. This choice has been made because the purpose of our algorithm is not to learn the basic skills required for driving, such as keeping the lane and following a trajectory, but it is instead choosing how to behave in traffic conditions and when to interact with other vehicles present in the environment. Moreover, a similar high-level perspective has also been implemented in other works, related to the centralised AIM paradigm \cite{klimke2022cooperative, klimke2022enhanced, guillenperez2022multiagent}.

Finally, for what concerns the reward function, we need to take into consideration the fact that each agent is trying to solve a multi-objective problem. Indeed the main goal of each vehicle is crossing the intersection and reaching the end of the scenario. In the meantime, a vehicle is also required not to collide with the others, by travelling as smoothly as possible. In order to fulfill all these objectives we design a reward signal composed by different terms:
\begin{equation} \label{eq:reward}
    r = \begin{cases}
        +x \hspace{0.85cm} &\text{if} \hspace{0.3cm} \mfa{x} > 0, \\
        -k &\text{if vehicle not moving,}\\
        -10\cdot k &\text{if a collision occurs,}\\
        +10\cdot k &\text{if scenario completed,}
    \end{cases}
\end{equation}
where $x\in \mathbb{R}_0^+$ is the distance travelled in meters from the previous time step and $k\in \mathbb{N}$ is a hyperparameter used to weight the importance of the last three components of the reward function with respect to the first one.

\subsection{Learning Strategy}
\label{subsec:learning}
The starting point for our learning strategy is the algorithm Deep Q-Networks, already presented in Section \ref{subsec:qldqn}. This algorithm is then slightly modified by considering the Double DQN scheme and also the Dueling architecture, which will be briefly introduced in the sequel. 

The idea of Double DQN \cite{van2016deep} is originated by the fact that Q-Learning, and consequently also DQN, are known to overestimate state-action values under certain conditions. This is due to the \textit{max} operation (see in (\ref{eq:updateql}) and (\ref{eq:loss})) performed to compute the temporal difference target. To mitigate this effect, the idea is to decouple the action selection and evaluation steps by using two different networks. We thus exploit the online network in the action selection step, while we keep using the target network for evaluation. This leads to the following modification of the loss function:
\begin{equation} \label{eq:lossdouble}
    \mathcal{L}(\theta) = \cfrac{1}{|\mathcal{B}|}\sum_{i\in \mathcal{B}} [(r_i+\gamma Q(s_i', \text{arg}\max_{a'}Q(s_i',a';\theta);\bar{\theta}) - Q(s_i,a_i;\theta))^2] .
\end{equation}
Dueling DQN \cite{wang2016dueling} instead introduces a modification in the NN architecture. Instead of having a unique final layer that outputs the Q-value for each possible action, we split it in two, with the first layer in charge of estimating the state value function (\ref{eq:vfunc}) and the second layer used for evaluating the advantage function (\ref{eq:advfunc}). These two quantities are then combined in the following way to produce an estimate of the state-action value function:
\begin{equation} \label{eq:dueling}
    Q(s,a;\theta,\alpha,\beta)=V(s;\theta,\alpha)+(A(s,a;\theta,\beta)-\cfrac{1}{|\mathcal{A}|}\sum_{a'}A(s,a';\theta,\beta)) ,
\end{equation}
where $\alpha$ and $\beta$ are the network parameters of the final layer, specific for the state-value function and advantage function respectively. Whereas, subtracting the term $\frac{1}{|\mathcal{A}|}\sum_{a'}A(s,a';\theta,\beta)$ is needed for stability reasons.

The final algorithm used for training is therefore a Multi-Agent version of Dueling Double DQN known as D3QN, with linearly-annealed $\varepsilon$-greedy policy for all the agents. In order to allow for decentralised execution while developing at the same time a smart training strategy, we consider an intermediate approach between the DTDE and the CTDE paradigms. In particular, we initialize and train three different D3QN agents, one for each turning intention, i.e. \textit{left}, \textit{straight} and \textit{right}. In this way each vehicle can select which model to use at the beginning of its path, according just to its own turning intention.

This approach is extremely sample-efficient, because we keep the number of network parameters constant, regardless of the number of vehicles considered. Moreover, these shared parameters are optimised through the experiences generated by all the vehicles, leading to a more diverse set of trajectories for training. Indeed, each of the three models has its own replay buffer, which contains transitions shared from all the vehicles with the corresponding turning intention. The crucial insight that makes our strategy effective is the fact that the observations gathered from each vehicle are invariant with respect to the road in which the vehicle itself is positioned. This \textit{parameter} and \textit{experience sharing} approach renders the training procedure of the algorithm somehow centralised because trajectories coming from different vehicles are used to train the three D3QN agents. However, we remark that, once the models have been trained, the execution phase is completely decentralised, since each vehicle locally stores the three different models. Then, at the beginning of the scenario, each CAV selects the model to use based only on its own turning intention.

\subsection{The prioritised scenario replay strategy}
\label{subsec:psr}
The agents are trained for a fixed number of iterations $N\in \mathbb{N}$, keeping the intersection busy in order to obtain meaningful transitions to learn from. In particular, at each episode we consider the most complicated situation in which there are four vehicles simultaneously crossing the intersection, one for each road and with random turning intention.

Every $E\in \mathbb{N}$ time steps we pause training and run an evaluation phase. During this period, the agents use a greedy policy to face all the possible scenarios described above. When the evaluation is completed, we use the inverse of the returns from all the scenarios to build a probability distribution, and in the following training window we sample the different scenarios according to such a distribution. In this way we allow the agents to learn more from the most complicated situations. We name this original training strategy \textit{prioritised scenario replay} because of its conceptual similarity with the \textit{prioritised experience replay} scheme \cite{schaul2015prioritized}, common in many RL algorithms. Algorithm \ref{alg:d3qn} illustrates the proposed learning strategy in detail.
\begin{algorithm} %\footnotesize
    \caption{MAD4QN-PS}\label{alg:d3qn}
    \begin{algorithmic}[1]
        \STATE Initialize three state-action value networks $Q_i$ with random parameters $\theta_i$, $i=1,2,3$
        \STATE Initialize three target state-action value networks $\bar{Q}_i$ with parameters $\bar \theta_i = \theta_i$, $i=1,2,3$
        \STATE Initialize three replay buffers $\mathcal{D}_i$, $i=1,2,3$
        \STATE Setup initial $\varepsilon$, decay factor $\varepsilon_d$, evaluation period $E$, target update period $\delta$, discount factor $\gamma$
        \STATE Uniformly initialize the scenarios probability distribution
        \STATE max\_episode\_steps $\leftarrow M$
        \STATE $n \leftarrow 0$
        \WHILE{$n < N$}
            \STATE episode\_terminated $\leftarrow$ False
            \STATE Randomly reset the environment
            \STATE episode\_steps $\leftarrow 0$
            \WHILE{\textbf{not} episode\_terminated}
                \STATE $V \leftarrow$ number of vehicles currently present in the simulation 
                \STATE Assign each vehicle to one of the three agents, based on its turning intention
                \STATE Collect observations for each vehicle $o_1^n,...,o_V^n$
                \FORALL{vehicles $v$ \textbf{in} $1,...,V$}
                    \STATE With probability $\varepsilon$ select a random action $a_v^n$
                    \STATE Otherwise $a_v^n \leftarrow \text{arg}\max_a Q_i(o_v^n,a;\theta_i)$, where $i$ depends on the turning intention of ~$v$
                \ENDFOR
                \STATE Apply actions $a_v^n$ and collect observations $o_v^{n+1}$ and rewards $r_v^n$ for $v$ in $1,...,V$
                \STATE Store each transition in the corresponding replay buffer $\mathcal{D}_i$, $i=1,2,3$
                \FORALL{agents $i=1,2,3$}
                    \STATE Sample random batch of transitions $\mathcal{B}_i$ from replay buffer $\mathcal{D}_i$
                    \STATE Update parameters $\theta_i$ by minimising the loss function:
                    \begin{equation*}
                        \mathcal{L}(\theta_i) = \cfrac{1}{|\mathcal{B}_i|}\sum_{b\in \mathcal{B}_i} [(r_b+\gamma \bar Q_i(o_b', \text{arg}\max_{a'}Q_i(o_b',a';\theta_i);\bar{\theta_i}) - Q_i(o_b,a_b;\theta_i))^2]
                    \end{equation*}
                \ENDFOR
                \STATE $\varepsilon \leftarrow \varepsilon - \varepsilon_d$
                \STATE episode\_steps $\leftarrow$ episode\_steps + 1
                \STATE $n \leftarrow n + 1$
                \IF{$n$ \% $\delta == 0$}
                    \STATE Update target network parameters $\Bar{\theta}_i = \theta_i$ for each agent $i=1,2,3$
                \ENDIF
                \IF{$n$ \% $E == 0$}
                    \STATE Run evaluation phase and update the scenarios probability distribution as described in Section \ref{subsec:psr}
                \ENDIF
                \IF{a collision occurred \textbf{or} $V == 0$ \textbf{or} episode\_steps == max\_episode\_steps}
                    \STATE episode\_terminated $\leftarrow$ True
                \ENDIF
            \ENDWHILE
        \ENDWHILE
        
    \end{algorithmic}
\end{algorithm}

\section{Experiments on virtual environments}
\label{sec:experiments}
In order to train and evaluate our algorithm we need a suitable simulation environment. For this project we have chosen the platform SMARTS \cite{zhou2021smarts}, explicitly designed for MARL experiments for autonomous driving. SMARTS relies on the external provider SUMO (Simulation of Urban MObility) \cite{SUMO2018}, which is a widely used microscopic traffic simulator, available under an open source license. For our setup, we have used SMARTS as a bridge between SUMO and the MARL framework, since it follows the standard Gymnasium APIs \cite{towers_gymnasium_2023}, widely used in the RL community.

To develop our code Python 3.8 was employed along with the version 1.4 of the Deep Learning library PyTorch \cite{paszke2017automatic}. Moreover, a NVIDIA TITAN Xp GPU was used to run our experiments.
As already mentioned in Section \ref{subsec:sysmod}, we have built a 4-way 1-lane intersection scenario, with three different turning intentions available to the vehicles coming from each of the four ways. The simulation step has been fixed to 100$ms$. Regarding the observation of each vehicle, we have stacked $n=3$ consecutive frames, each consisting of a RGB image of dimensions $48 \times 48$ pixels. Whereas, the action space contains $m=2$ possible velocity commands\footnote{\mtc{Such a choice for the action space has been made because we observed that considering more velocity commands only introduced more complexity in the system, without increasing the performance of the algorithm.}}, namely $0$ and $15$$m$/$s$. 
\mtc{The chosen velocity references} are then fed at each iteration to a speed controller, in charge of driving the vehicle until the subsequent time step. For what concerns the reward function (\ref{eq:reward}), we have fixed its hyperparameter to $k=1$. Regarding the architecture of the NN used to approximate the state-action value function, we have considered a convolutional neural network (CNN), whose structural details are summarised in Table~\ref{tab:cnn}. Finally, the training hyperparameters of Algorithm~\ref{alg:d3qn} are reported in Table~\ref{tab:params}.

\begin{table*}[t]
  \caption{CNN architecture}
  \label{tab:cnn}
  \small 
  \centering
  \begin{tabular}{lccccc}
    \toprule
    Layer&N. of Filters&Kernel Size&Stride&Activation Function&N. of Neurons\\
    \midrule
    Convolutional & $32$ & $8\times 8$ & 4 & ReLU & / \\
    Convolutional & $64$ & $4\times 4$ & 2 & ReLU & / \\
    Convolutional & $64$ & $3\times 3$ & 1 & ReLU & / \\
    Fully connected & / & / & / & ReLU & 512 \\
    Fully connected (V) & / & / & / & Linear & 1 \\
    Fully connected (A) & / & / & / & Linear & 2 \\
    \bottomrule
  \end{tabular}
\end{table*}

\begin{table*}[t]
	\caption{Training hyperparameters}
	\label{tab:params}
    \small 
    \centering
	\begin{tabular}{lcl|lcl}
		\toprule
		Hyperparameter & Symbol & Value & Hyperparameter & Symbol & Value\\
		\midrule
		Training steps & $N$ & $10^6$ & Initial exploration rate  & $\varepsilon$ & $1$\\
		Max episode steps & $M$ & $10^3$ & Exploration rate decay &  $\varepsilon_d$ & $10^{-6}$\\
		Evaluation period & $E$ & $5\cdot 10^3$ & Buffer size & $|\mathcal{D}|$ & $1.5\cdot 10^5$\\
		Optimizer & / & RMSprop \cite{hinton2012lecture} & Discount factor & $\gamma$ & $0.99$ \\
		Learning rate & $lr$ & $10^{-4}$ & Batch size & $|\mathcal{B}|$ & $256$ \\
		Target update period & $\delta$ & $10^3$ & & & \\
		\bottomrule
	\end{tabular}
\end{table*}

\subsection{Baselines}
\label{subsec:baselines}
In order to assess the quality of our algorithm, we benchmark it versus the following baselines~\footnote{The baselines simulations with traffic lights have been performed exploiting Flow \cite{wu2021flow}, another platform used to interface with SUMO, which easily allows for the definition and control of traffic lights.}:
\begin{itemize}
    \item \textit{Random policy} (RP) for all the vehicles, which helps confirm whether our algorithm is effectively learning meaningful patterns, as it demonstrates its ability to outperform random actions, which lack any deliberate learning process.
    \item \mtc{Three symmetric (N/S \& W/E) \textit{fixed-time traffic lights} (FTTL), considering two cycle lengths already analyzed in \cite{klimke2022enhanced, guillenperez2022multiagent}, and also the optimal cycle length in function of the traffic flow, computed according to Webster's formula \cite{webster1958traffic}. The final flow rate of vehicles for evaluation has been set to 600 veh/hour, as will be discussed in Section \ref{subsec:res}.}
    \item Two symmetric (N/S \& W/E) \textit{actuated traffic lights} (ATL) \cite{SUMO2018}, with different cycle lengths, which operate by transitioning to the next phase once they identify a pre-specified time gap between consecutive vehicles. In this way the allocation of green time across phases is optimised and the cycle duration is adjusted in accordance with changing traffic dynamics. 
\end{itemize}
The parameters of the \mtc{five} traffic lights configurations are reported in Table \ref{tab:tl}.
\begin{table*}[t]
  \caption{Traffic lights parameters}
  \label{tab:tl}
  \centering
  \small 
  \begin{tabular}{lcccc}
    \toprule
    Traffic light&Red \& Green&Yellow&Min. Duration&Max. Duration\\
    \midrule
    FTTL1 & $25s$ & $5s$ & / & / \\
    FTTL2 & $32s$ & $8s$ & / & / \\
    FTTLOPT & $15s$ & $2s$ & / & / \\
    ATL1 & $25s$ & $5s$ & $10s$ & $40s$ \\
    ATL2 & $32s$ & $8s$ & $15s$ & $50s$ \\
    \bottomrule
  \end{tabular}
\end{table*}
\\\\
As a final note, we emphasize that we do not have considered any RL-based centralised AIM approach as baseline because the purpose of our method is to propose a more realistic and feasible alternative to them, which is however able to outperform classical intersection control methods, such as traffic lights.

\subsection{Results}
\label{subsec:res}
To assess the quality of MAD4QN-PS we consider four metrics, namely the \textit{travel time}, the \textit{waiting time}, the \textit{average speed} and the \textit{collision rate}. We remark that we have accounted for vehicle-centered metrics because of the decentralised nature of our algorithm. However, it is evident that by optimizing the performance of each single road user we also implicitly improve the quality of the whole intersection. 
The robustness of our method is ensured by performing training ten times across different seeds, and then considering all the different trained models while evaluating our strategy. In particular, each model has been tested by running a cycle of the evaluation phase presented in Section \ref{subsec:psr}\mtc{, considering 600 veh/hour as flow rate of vehicles coming through the intersection. Then,} the obtained results from the different models have been averaged out. Moreover, to ensure a fair comparison and analysis of the results, the same evaluation setup has been adopted for all the baselines introduced above. \mtc{Finally, it is worth noticing that the inference time of the networks at evaluation phase is at most 1$ms$, thus allowing for real-time control, given that the simulation step has been fixed to 100$ms$, as discussed at the beginning of this section.}

\begin{figure}[t]
	\centering
	\begin{subfigure}[b]{0.40\textwidth}
		\includegraphics[width=\textwidth]{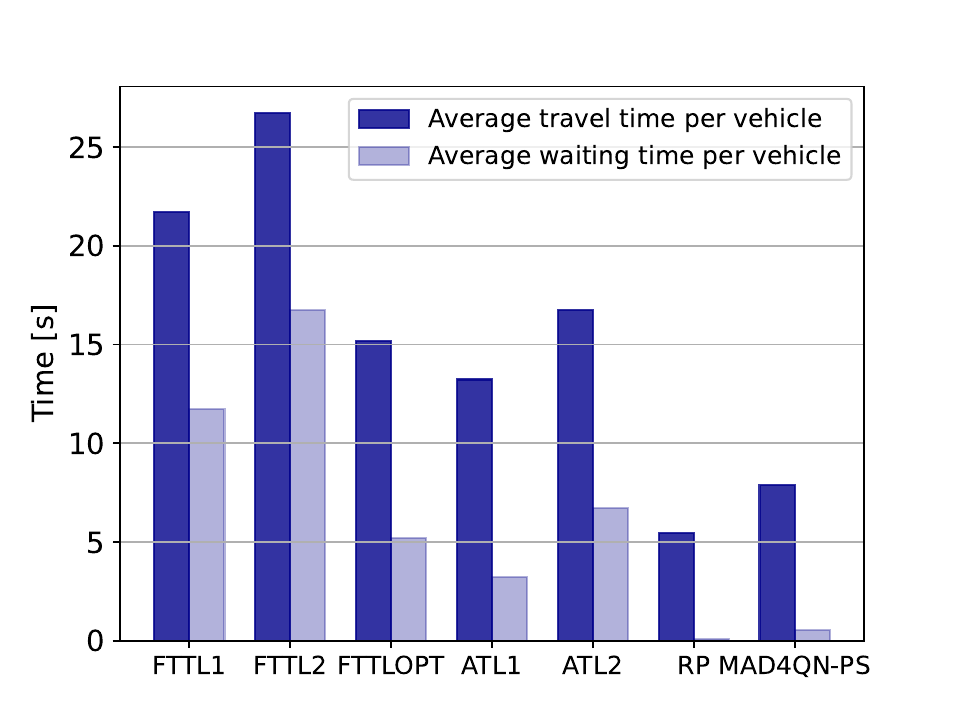}
		\caption{Average travel and waiting time per vehicle}
		\label{fig:twtime}
	\end{subfigure}
	%\hfill
	\begin{subfigure}[b]{0.40\textwidth}
		\includegraphics[width=\textwidth]{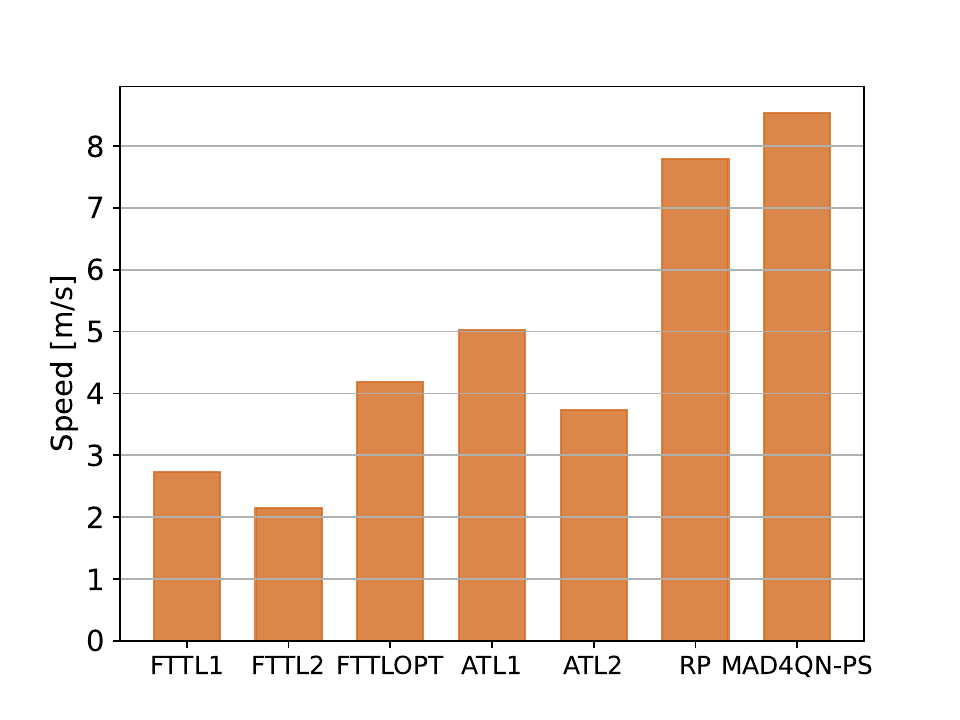}
		\caption{Average speed per vehicle}
		\label{fig:avgspeed}
	\end{subfigure}
	%\hfill
	\begin{subfigure}[b]{0.40\textwidth}
		\includegraphics[width=\textwidth]{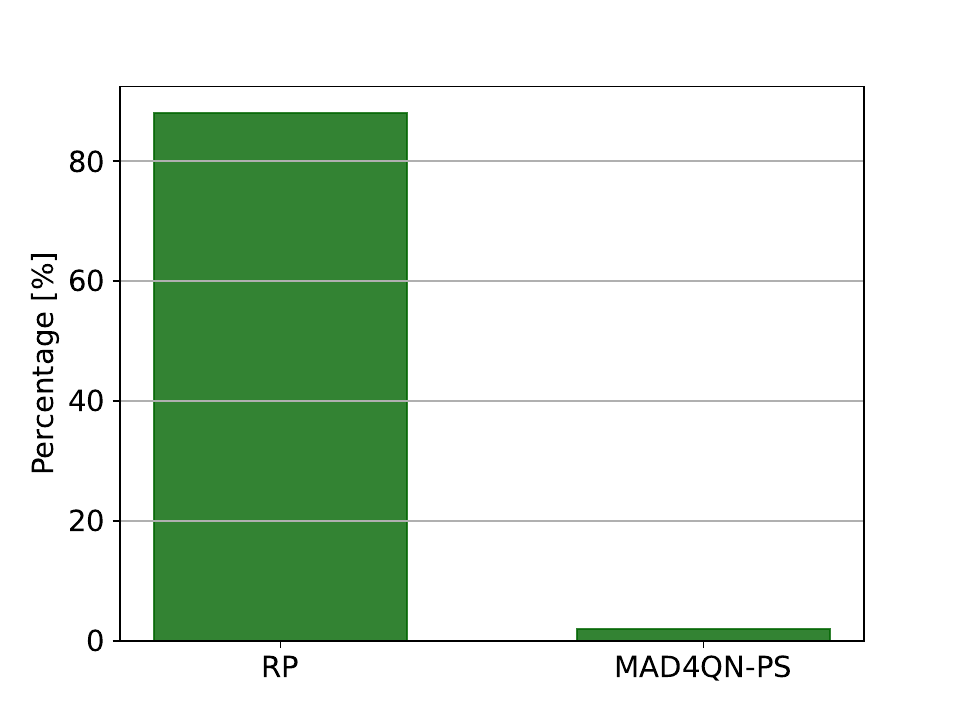}
		\caption{Average collision rate}
		\label{fig:collisions}
	\end{subfigure}
	\caption{Comparison between the performance metrics of our method (MAD4QN-PS) and the baselines introduced in Section \ref{subsec:baselines}.}
	\label{fig:results}
\end{figure}

Starting from Figure \ref{fig:twtime}, we can observe the average travel time and waiting time of a generic vehicle for all the methods. The former is defined as the overall time that the vehicle spends inside the environment, while the latter is defined as the fraction of the travel time in which the vehicle is moving with velocity less or equal to 0.1$m$/$s$, i.e. when it is stopped or almost stopped. We clearly see that our method strongly outperforms all the traffic lights configurations. This is mainly due to the fact that, when using traffic lights, a fraction of the vehicles is forced to stop as soon as the corresponding light becomes red. Conversely, the trained MAD4QN-PS agents are able to smoothly handle the interaction among multiple vehicles, allowing them to avoid stopping unless it is strictly necessary.
Figure \ref{fig:avgspeed} instead displays the average speed of each vehicle. The results shown in this histogram are clearly related to those in Figure \ref{fig:twtime}; indeed, also in this case we can see that our method outperforms traffic lights control schemes. This occurs since the vehicles almost never stop, thus keeping a smoother velocity profile throughout all the duration of the simulation.

% 0.45 before

Lastly, we are left with the analysis of the random policy baseline, as we need to look at all the three plots to fully understand its behaviour. If we just look at Figures \ref{fig:twtime} and \ref{fig:avgspeed} we could argue that the random policy performance is similar to that of MAD4QN-PS. This hypothesis is however disproved by Figure \ref{fig:collisions}, where the average collision rate for each vehicle is illustrated. The extremely high collision percentage obtained by the random policy explains why each vehicle on average spends a small amount of time with high velocity into the environment. Indeed, the simulation is terminated as soon as a vehicle crashes. MAD4QN-PS, instead, achieves an extremely low collision rate. In addition, the fact that such a collision rate is non-zero is expected and also observed in other works exploiting RL-based techniques  \cite{klimke2022cooperative, klimke2022enhanced}, given that our algorithm has to implicitly learn collision avoidance through the reward signal.
%%%
In practice, the remaining failures are not problematic, because we can integrate rule-based sanity checks in the pipeline in order to be 100\% collision-free.
Additionally, we note that two out of the ten trained models with different seeds achieve exactly $0\%$ collision rate, meaning that if we select one of those models for deployment we are able to attain collision-free performances. This is interesting since from an applicability perspective only the best trained model would be used in practice.
%%%
As a final note, we have not plotted the collision rate of the traffic lights methods for better visualization, since the latter quantity is trivially zero for all the configurations.

A short video showing the performances of MAD4QN-PS can be found at \href{https://youtu.be/nDhdnyNLlbM}{this link}.

\section{Conclusions and future directions}
\label{sec:conclusions}
In this study, we consider a distributed approach to face the AIM paradigm. In particular, we propose a novel algorithm which exploits MARL through self-play and an original learning strategy, named \textit{prioritised scenario replay}, to train three different intersection crossing agents. The derived models are stored inside CAVs, that are then able to complete their paths by choosing the model corresponding to their own turning intention while relying just on local observations.
Our algorithm represents a feasible and realistic alternative to the centralised AIM concept, that is still expected to require years of technological advancement to be implementable in a real-world scenario. In addition, simulation experiments demonstrates the superior performances of our method w.r.t. classic intersection control schemes, such as static and actuated traffic lights, in terms of travel time, waiting time and average speed for each vehicle.

In future works, we aim to explore different directions for advancements. In particular, one of the main objectives is to also consider human driven vehicles inside the environment \mfa{and extend our approach to this field of research (see, e.g. the initial effort made in \cite{yan2021reinforcement}). In this case, the most challenging issue is indeed represented by the synchronization of traffic lights accounting for the presence of human driven vehicles. Moreover,} given the decentralised nature of the proposed method, we expect to render our algorithm more robust without dramatically change it. Conversely, significant redesign would be necessary for a centralised AIM approach. %, but we will still need to make some adjustments in order to render our models more robust. 
Furthermore, we envisage to test more complicated scenarios, both in terms of dimension and layout, again to improve the robustness of our algorithm. Finally, we intend to implement our algorithm in a scaled real-world scenario with miniature vehicles \cite{paull2017duckietown}, to practically demonstrate the applicability of our method.

\bibliography{references}

\end{document}